\documentclass[11pt]{article}
\usepackage{times}
\usepackage{amsmath,amsthm,amssymb,setspace,enumitem,epsfig,titlesec,verbatim,color,array,multirow,comment,graphicx,hyperref}
\graphicspath{ {./images/} }

\usepackage[super,sort&compress,comma]{natbib}
\usepackage[bf,small]{caption}
\usepackage[export]{adjustbox}
\usepackage[top=2.5cm,left=2.6cm,right=2.6cm,bottom=3.5cm]{geometry}
\usepackage{bibunits}
\usepackage{multicol}
\usepackage{amsmath}
\usepackage{xcolor}
\usepackage{dcolumn }
\usepackage{lineno}
\usepackage{booktabs}
\usepackage{microtype}
\smallskip

\makeatletter
\providecommand{\doi}[1]{}
\renewcommand{\doi}[1]{}

\renewcommand{\url}[1]{\unskip\@gobble}
\makeatother

\titleformat{\section}{\sffamily \fontsize{13}{20}\bfseries}{\thesection}{1em}{}
\titleformat{\subsection}{\sffamily \fontsize{12}{20}\bfseries}{\thesubsection}{1em}{}
\titleformat{\subsubsection}{\sffamily \fontsize{11}{20}\bfseries}{\thesubsubsection}{1em}{}
\makeatletter
   \def\tagform@#1{\maketag@@@{[#1]\@@italiccorr}}
\makeatother

\title{\sffamily \Large {\bfseries LLMs struggle to simulate human belief updates in controlled environments}}
\date{\empty}
\author{\parbox[c]{16cm}{\centering \onehalfspacing Sebastian Pohl$^{1,*}$, Harsh Mehta$^1$, Pranav Mambayil$^1$, Abdul Ghafoor$^1$, Franziska Lesigang$^1$, Yufang Hou$^{1,\dag}$, Christian Hilbe$^{1,\dag}$\\[0.2cm]
		$^1$IT:U, Interdisciplinary Transformation University Austria,\\ Altenberger Straße 66c, 4040 Linz, Austria\\
		$^\dag$ Equal supervision.\\
        $^*$ Corresponding author: sebastian.pohl@it-u.at\\[1cm]
		}}

\begin{document}
\maketitle
\onehalfspacing


\section*{Abstract}

LLMs are increasingly deployed as proxies for human study participants in social science experiments, yet the fidelity of this practice has rarely been tested directly. We test whether six LLMs can simulate individual human belief updates, comparing LLM outputs 1-to-1 against ground truth data from 391 UK participants on Prolific, who updated their stances on three discussion topics after reading Reddit comments. Each participant was simulated by an LLM conditioned on a persona derived from their demographic and personality trait data.
Given participants' actual initial stances, five of the six models showed no systematic directional bias in their simulated post-stances, although the size of their individual-level errors varied considerably between models.
All six models fail to simulate initial stances themselves and to produce faithful belief updates from self-generated stances. Three systematic biases emerge across all models: overrepresentation of neutral positions, more frequent but smaller belief shifts than humans, and a failure to rank comments by convincingness. Demographic and personality trait personas had no consistent effect on fidelity. We conclude that without grounding in realistic initial stances, LLM simulations of human belief dynamics cannot be considered reliable, a condition that current multi-round social media simulations rarely meet.


\newpage
\section*{Introduction}
Simulating human behavior with large language models (LLMs) has emerged as a promising new tool in the scientific workflow \cite{gao2024large}. Deploying LLMs as stand-ins for human subjects in controlled experimental settings has the promise to enable researchers to test initial hypotheses at lower cost and with greater speed than traditional human studies \cite{park2023generative, YaxAnlloPalminteri2024Reasoning, gao2024large, Wang2024asurvey, gao2025takecaution, cui2025largescale, Sakamoto2025value, wang-etal-2025-decoding, tang-etal-2025-gensim, piao2025agentsocietylargescalesimulationllmdriven, zeng2026toohuman}.
The field of social media debate dynamics has recently attracted a growing number of studies, where LLMs are used as proxies for human study subjects \cite{chuang-etal-2024-simulating, gao2024large, wang-etal-2025-decoding, piao2025emergencehumanlikepolarizationlarge, piao2025agentsocietylargescalesimulationllmdriven}. LLM simulations may also enable interventions on systems that would otherwise be inaccessible for practical or ethical reasons, for example, when studying polarization dynamics or how information spreads through social networks \cite{chuang-etal-2024-simulating, gao2024large, Wang2024asurvey, wang-etal-2025-decoding}. A related line of work investigates how LLM simulations give rise to collective behaviors, such as the emergence of social norms, beyond replicating the behavior of individual humans \cite{park2023generative, ren2024emergence}. Beyond hypothesis testing, reliable human-like LLM simulations have further applications in benchmarking LLM agents, where realistic user simulators enable researchers to test agent behavior in more applicable and generalizable environments \cite{Zhu-2024-how, wang2025user, bougie2025simuser, yao2025taubench, barres2025tau2benchevaluatingconversationalagents} and in supporting skill development through simulated social interactions \cite{yan2025social}. At the same time, as LLMs increasingly act within hybrid human-AI systems, understanding how their behavior compares to that of humans matters well beyond simulation studies \cite{han2026socialphysicsageartificial}.

The same human-likeness that makes LLMs attractive as simulators also carries risks. Content produced by LLMs that is increasingly hard to distinguish from human-created content may be used to spread misinformation \cite{pan-etal-2023-risk, chen2024can}. In research, the presence of LLM pollution, where responses assumed to be human are partly or wholly produced by LLMs, threatens the validity of studies that draw on internet sources to study human behavior \cite{rilla2025recognisinganticipatingmitigatingllm}. Given both the opportunities and the risks, it is important to gain a better understanding of how human-like LLMs are.

\begin{figure}[t]
    \centering
    \includegraphics[width=\linewidth]{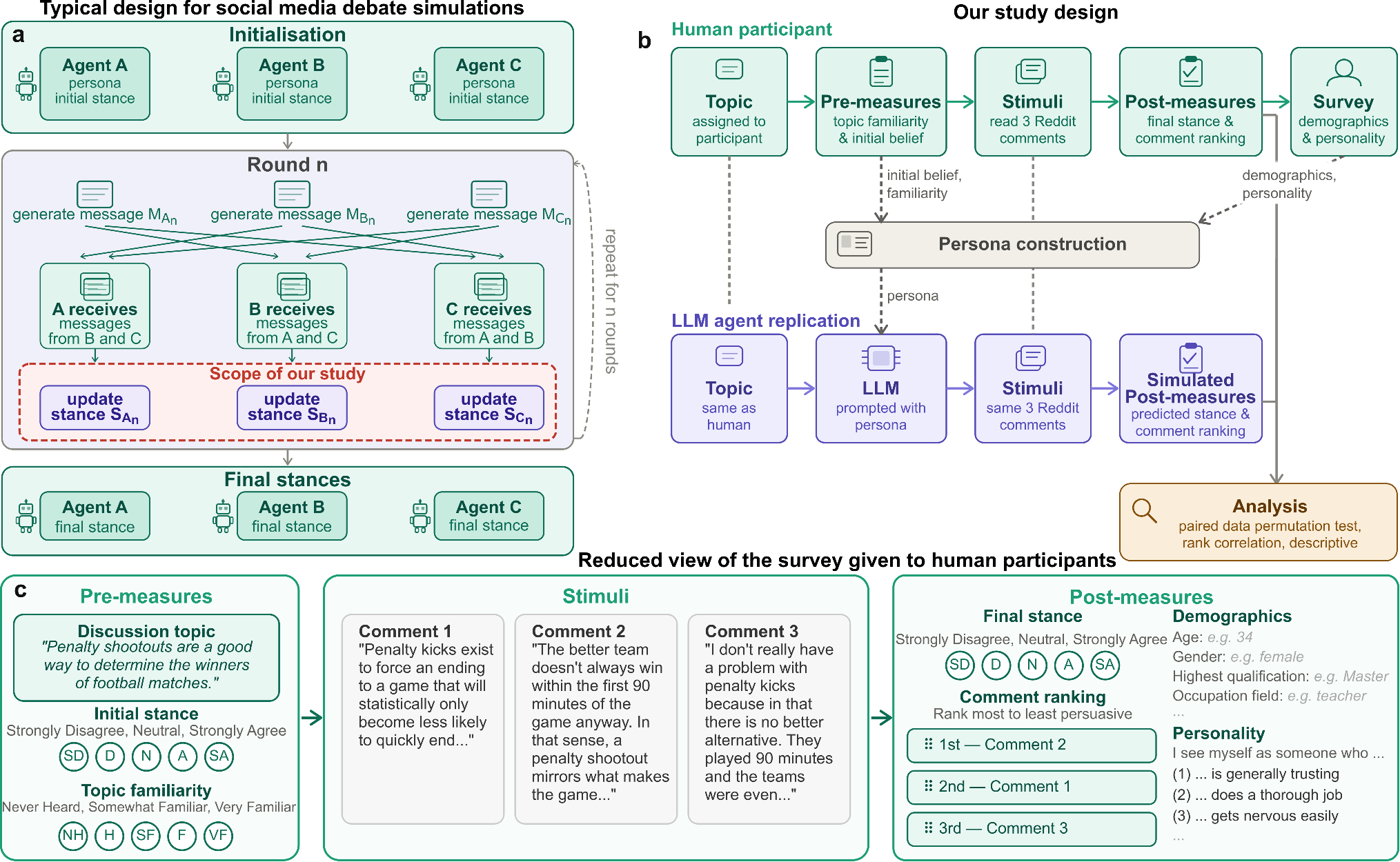}
    \caption{\textbf{Overview of the design of social media debate simulations and our contribution.} Panel \textbf{a} shows the typical design of social media debate simulations. Panel \textbf{b} zooms in to the belief update step within that design that we target in this study. Panel \textbf{c} zooms in to the human data collection aspect of panel b, describing the full survey (pre-measures, post-measures, and final survey) given to human participants. \textbf{a}, This study targets the highlighted (red) belief update stage that closes each communication round before agents generate new messages for the next round based on their updated stances. \textbf{b}, For our study, we collect ground truth data from participants on Prolific that update their stances on a discussion topic after reading three reddit comments. These human belief updates are then simulated by LLMs 1-to-1, where we construct a tailored persona for each study participant and expose the LLM to the exact same information that participant saw. \textbf{c}, Each participant completes the pre-measures, stimuli and post-measures (final stance and comment rankings) for all three topics before they finally fill out the demographics and personality trait survey.}
    \label{fig:study_overview}
\end{figure}

In this work, we investigate this question in the context of social media debate simulations, focusing on the viability of LLMs as stand-ins for human participants. A typical simulation in this domain follows the structure outlined in Figure~\ref{fig:study_overview}a. LLM agents alternate between generating and exchanging messages on some discussion topic with each other and updating their beliefs on the topic over several communication rounds to form a simulated social network \cite{chuang-etal-2024-simulating, gao2024large, wang-etal-2025-decoding, piao2025emergencehumanlikepolarizationlarge, piao2025agentsocietylargescalesimulationllmdriven}.

Existing validations of social media debate simulations do not directly compare individual LLM agents against matched human participants. Most studies assess simulation outcomes at the aggregate level, matching results against mathematical models (e.g., Friedkin-Johnsen \cite{FriedkinJohnsen1990}, Bounded Confidence \cite{DeffuantEtAl2000MixingBeliefs}, or agent-based models \cite{gilbert2000agent}), against polarization patterns in humans \cite{wang-etal-2025-decoding, chuang-etal-2024-simulating}, or against real-world social media engagement counts \cite{zhang-etal-2025-ga}. In none of these cases are LLM and human data collected under strictly comparable conditions. Direct comparisons between humans and LLMs do exist in other domains, for example examining trust behavior or reasoning depth in game-theoretic settings \cite{xie2024can, gao2025takecaution} and the replication of moral judgments \cite{GrizzardEtAl2025ChatGPTMoralJudgments}. But they follow a different experimental design than what is used in the domain of social media debate dynamics. Controlled replications of psychology and management experiments \cite{cui2025largescale} or of studies on reliance on human versus AI advice \cite{rieger2026ChatGPT} are more closely related in topic, but focus on aggregate and interaction effects rather than 1-to-1 participant matching.

We isolate the belief update step of social media debate simulations and test whether six LLMs of varying size and release date can simulate human belief updates in a controlled setting. We additionally test 18 models to examine how model size and post-training stage affect fidelity. We address a crucial gap in existing evaluations of social media debate simulations, which only validate the outcome of these simulations. Instead we break down the multi-turn communication setup of previous studies to ask whether individual simulation steps are sound on their own. As illustrated in Figure~\ref{fig:study_overview}a (red dashed box), we target the belief update step, where an LLM agent is presented with messages from their peers and has to update their beliefs in a human like manner. The simulation would then proceed with each agent generating new messages based on their updated beliefs to initiate a new communication round. We argue that in particular simulations used to evaluate interventions, for example, whether active or passive nudging strategies can reduce polarization in social networks \cite{wang-etal-2025-decoding}, should be assessed not only by whether they reproduce human outcome distributions, but also by whether they capture the underlying belief-update processes that generate these outcomes. If the internal dynamics differ substantially from those of human participants, conclusions about the effectiveness of such interventions may not transfer reliably from LLM agents to real-world settings.

Focusing on a single step also makes a controlled experimental design possible. Unlike previous validations, this allows a direct comparison between each human participant and the LLM agent simulating them. In our experiment, we recorded data from 391 participants on Prolific. Given a discussion topic and a number of social media comments sourced from the \textit{r/changemyview} subreddit \cite{reddit_cmv}, we recorded how humans changed their beliefs after reading these comments and tested whether LLMs update their beliefs in the same way that humans do. LLMs were instructed to replicate human subjects with personas that contain key demographic information and personality traits. Figure~\ref{fig:study_overview} illustrates our experimental design. Crucially, we have a paired design, where for each human participant, we set up a direct simulation with a tailored LLM agent based on their demographic and personality trait data. We use this setup to address three main research questions. \textbf{(Q1)} Do LLMs accurately simulate belief changes recorded from our human participants? \textbf{(Q2)} Are there systematic differences between the update patterns between humans and LLMs? \textbf{(Q3)} What role does the persona, the LLM is conditioned on, play for the accuracy of the simulation?

Our experiments paint a mixed picture of the ability of LLMs to simulate human belief updates. A paired test on post-stances detected a systematic directional bias only for Llama-3.3-70B-Instruct among the six main models, yet the remaining models still differed considerably in their individual-level error, measured as the mean absolute error (MAE) between simulated and human post-stances. Newer and more capable LLMs did not always perform better on this simulation task, although within a model family, larger models produced a smaller MAE. Across all LLMs, we further found three systematic differences between human and simulated belief updates. First, LLMs showed a bias towards neutral positions and underestimated the frequency of extreme opinions. Second, LLMs updated their beliefs more frequently than humans, but with lower magnitude. And third, LLMs failed to predict how convincing humans found individual comments. Importantly, simulation fidelity hinged on the starting point. Without access to participants' initial beliefs, LLMs failed to predict post-stances altogether. The demographic and personality personas commonly used to condition LLMs in such studies \cite{chuang-etal-2024-simulating, xie2024can, wang-etal-2025-decoding, Zhu-2024-how, wang2025user} did not enable them to generate human-like initial beliefs. Only when each agent had access to the initial belief of its human counterpart did the simulations show promise.

Our results thus suggest that, if LLMs can simulate human belief updates at all, they can do so only under specific conditions. Unlike previous work, which reports increased polarization or amplified effects \cite{GrizzardEtAl2025ChatGPTMoralJudgments, wang-etal-2025-decoding, Acrebi2023large}, we observe a tendency to regress towards neutral responses. More broadly, simulation fidelity depends on grounding models in context-specific empirical data, not on demographic or personality conditioning alone. This challenges researchers to identify the task-relevant features that are essential for accurate simulation in their setting, which in our case are in particular initial beliefs. It also points to a systematic risk we term \textit{simulation drift}. As a multi-round simulation progresses, initial belief distributions in each belief update step drift away even from a human ground truth starting point. Given the critical importance of accurate initial beliefs for simulation fidelity, the simulation risks drifting outside the regime in which LLMs can still faithfully reproduce human belief updates.

Along with this paper, we also release our dataset to support future research, enabling others to evaluate how well newly developed agents can simulate the belief updates we recorded in this controlled setting.

\section*{Results}
Our experiments implement the belief update step present in typical simulations of social media debate dynamics \cite{chuang-etal-2024-simulating, gao2024large, wang-etal-2025-decoding, piao2025emergencehumanlikepolarizationlarge, piao2025agentsocietylargescalesimulationllmdriven}. The experimental design is illustrated in Figure~\ref{fig:study_overview}. First, we asked our 391 human participants to rate their initial stance on and familiarity with a given discussion topic \textit{(support of universal basic income, penalty shootouts to decide football matches, or use of weight loss drugs like Ozempic)}. Chosen at random, half of the participants received a negated version of each topic statement, so that their agreement ratings map to disagreement with the original framing and vice versa. The Supplementary Information (Supplementary Table~5) gives additional information on the effects of topic framing. Then participants were presented with three messages from their peers. In our case, we hand selected three message packages per topic from the \textit{r/changemyview} subreddit \cite{reddit_cmv} with an overall more pro, contra or neutral stance. After reading these messages, participants were asked again to rate their current stance on the discussion topic and to rate the three comments they read by how convincing they found them. This was repeated three times for the three topics before we finally asked the participants to complete a short demographic and personality trait survey. Figure~\ref{fig:study_overview}c illustrates the survey given to human participants.

Second, we set up a direct replication of this data collection process using LLM agents. For each of the human participants, we constructed a tailored persona, including their topic familiarity, initial stance and demographic and personality trait survey responses. Consequently, we have a paired design, where each human participant is simulated by an LLM agent that is instructed to simulate the given humans post comment measures \textit{(post stance and comment ranking)}, having access to the same information that the participant was presented with as well as their persona profile. Further details of our experimental design and data collection process are given in the methods section.

\begin{figure}[!htbp]
    \centering
    \includegraphics[width=\linewidth]{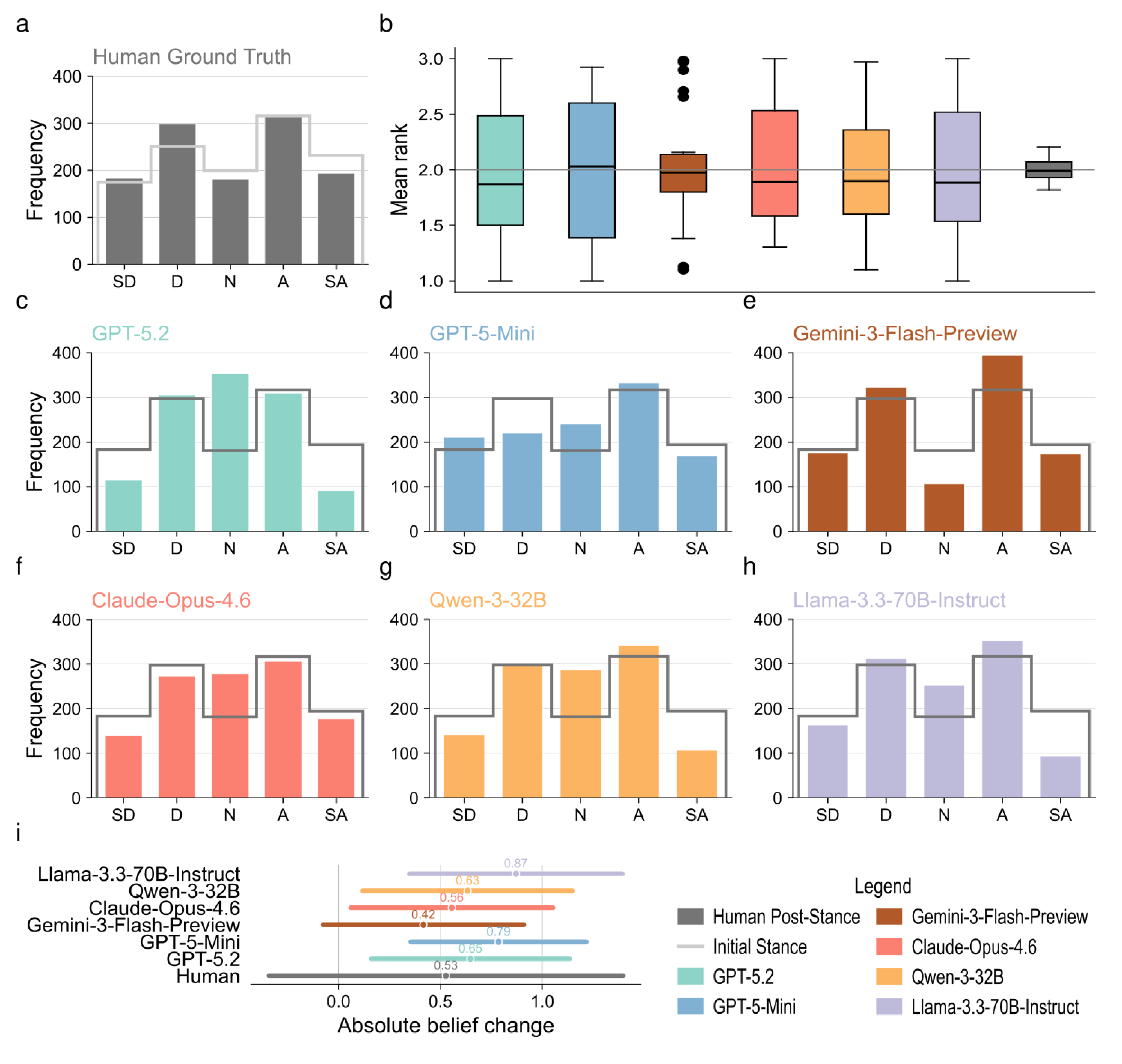}
    \caption{\textbf{Results of our main experiments comparing human to simulated belief updates.} SD=strongly disagree, D=disagree, N=neutral, A=agree, SA=strongly agree. \textbf{a}, Aggregated across topics the human post-stance distribution shifted slightly more towards disagree compared to the human initial stances (gray outline). \textbf{b}, The boxplot (median and percentiles) of the average ranks assigned to the 27 comments (9 comments per 3 topics) by all rankers shows that humans disagree with each other more on comment rankings than LLMs. For a given comment some participants would rank it as most convincing, whereas others would rank it last. As a result the average ranks of comments are grouped more closely around the mean. LLMs more consistently rank individual comments as convincing or unconvincing than humans, resulting in more extreme average ranks for these comments. \textbf{c-h}, The simulated post stance distributions plotted against an outline showing the human post stance distribution reveal that LLMs tend to concentrate less weight on the extreme positions. \textbf{i}, The mean and standard deviation of the absolute stance deltas (post - initial stance) for humans and LLMs indicate that LLMs change their beliefs more frequently but with lower magnitude. They have a slightly higher average delta, but lower standard deviation than humans.}
    \label{fig:results_1_overview}
\end{figure}

Figure~\ref{fig:results_1_overview}a shows the initial and post-stances recorded from our human study participants aggregated across all three topics. The overall distribution did not change greatly, though comments moved participants slightly toward more balanced positions, on average. More detailed information about the directions of change in human stances can be found in Supplementary Figures~7 and~8 that show the differences between post and initial stances split by discussion topic and initial belief.

\subsection*{Comparison of paired belief changes}
Our first question \textbf{(Q1)} is whether LLMs accurately simulate belief changes recorded from our human participants. To address this, we leverage our paired study design and apply a two-sided permutation test on paired stance differences, thus assessing simulation fidelity not just at the distribution level, but at the level of individual participants. Since positive and negative errors can cancel each other out in the mean paired stance difference, the permutation test is primarily sensitive to directional bias in updates. We additionally report the mean absolute error (MAE) for each model. This tests our null hypothesis (\(H_0\)) that \textit{LLMs produced faithful belief updates corresponding to human ground truth data} against our alternative hypothesis (\(H_1\)) that \textit{LLMs simulate belief updates that systematically differ from the belief updates recorded from humans}.

\begin{table}[ht] \centering
\caption{The table illustrates how valid the models are at simulating the subjects. The mean paired final stance difference is the average of human post-stance minus matched LLM post-stance across all participants; a value close to zero indicates that the model's simulated stances align well with human stances on average. The p-value is the probability of a mean paired difference at least as extreme as the observed one if LLM and human stances did not differ systematically. It is only sensitive to a directional bias in the data. Since positive and negative errors can cancel each other out a simulation can still make a large, but random error that does not show up in the mean paired final stance difference or p-value. Hence we also report the mean absolute error (MAE) for each model. Bold values are significant.}\label{tab:main_permutation_test}
\begin{tabular}{l c c c}
\toprule
Model & Mean Paired Final Stance Difference & p-Value & MAE\\
\midrule
Llama-3.3-70B-Instruct & 0.1185 & $<$\textbf{0.001} & 0.9027 \\
GPT-5.2 & 0.0725 & $0.0170$ & 0.7261 \\
Claude-Opus-4.6 & -0.0580 & $0.0582$ & 0.6775 \\
Qwen3-32B & 0.0554 & $0.0821$ & 0.7406 \\
Gemini-3-Flash-Preview & -0.0205 & $0.5196$ & 0.6433 \\
GPT-5-Mini & 0.0111 & $0.7524$ & 0.8191 \\
\bottomrule
\end{tabular}
\end{table}

With a Bonferroni correction for independent tests on six LLMs we set a per-test significance threshold of $\alpha_{\text{corrected}} = \frac{0.05}{6} \approx 0.0083$, which controls the familywise error rate at 5\%. As the results in Table~\ref{tab:main_permutation_test} show, for Llama-3.3-70B-Instruct, we can confidently reject \(H_0\). The belief updates produced by this model clearly differ from the human ground truth data. GPT-5.2 ($p = 0.0170$) falls below the uncorrected threshold but above the corrected one and is therefore not considered significant here. For other models (Claude-Opus-4.6, Qwen3-32B, Gemini-3-Flash-Preview and GPT-5-Mini), we cannot reject the null hypothesis either.

Further tests that split the data by topic and by message package show differences between human and LLM belief updates that cancel each other out in the aggregate across topics. The full results are reported in Supplementary Tables 1,2 and Supplementary Figure 1. In the aggregate, however, the main tests did not detect a systematic difference between the belief updates produced by these models and the human data. The models for which no difference was detected span a range of sizes and general capabilities, from Qwen3-32B and GPT-5-Mini to Claude-Opus-4.6 and Gemini-3-Flash-Preview.

The absence of a systematic difference does not mean that these models simulate individual participants equally well. A model can produce errors in both directions without shifting the mean. Table~\ref{tab:main_permutation_test} reports each model's MAE. We also compared the models directly with a permutation test on their mean absolute error (MAE) relative to the human post-stances (Supplementary Table~3). 13 of the 15 pairwise comparisons are significant after Bonferroni correction, so the models can be ranked by their MAE. This ordering does not simply follow model size or recency. Gemini-3-Flash-Preview, a lightweight variant in its model family, produces no larger MAE than the frontier model Claude-Opus-4.6 and a much smaller MAE than GPT-5.2. Qwen3-32B has a much smaller MAE than the larger Llama-3.3-70B-Instruct and performs on par with GPT-5.2. Within the GPT-5 family, the larger GPT-5.2 does produce a smaller MAE than GPT-5-Mini, so size is not irrelevant, but it does not predict on its own how well a model simulates human belief updates.

MAE and mean paired final stance difference do not always agree. Llama-3.3-70B-Instruct, for which we reject \(H_0\), also produces the largest MAE. Other models, however, for which the main test showed no significant difference, still produce large MAEs, in particular GPT-5-Mini, which showed the least mean paired final stance difference. Although the post-stances predicted by GPT-5-Mini often differed from the actual human post-stances, there was no systematic bias to these differences that would have shown up in the paired permutation test on mean paired final stance differences. In the following, we report results that highlight some of these additional differences between human and LLM belief updates.



\subsection*{Systematic differences between humans and LLMs}
We also looked at whether LLMs can accurately simulate the belief changes recorded from human participants at the distribution level. To this end we used a non-directional Chi-squared test of independence to compare human and LLM post-stances. At the distribution level we find all LLMs to produce post stance distributions that are significantly different from the human ground truth, with $p<0.0001$ for all LLMs. The Chi-squared test is more sensitive to differences in variances between distributions than the paired permutation test when the means are close in values, which is reflected here. Looking at the distributions in Figure~\ref{fig:results_1_overview}c-h, we can observe in response to our second question \textbf{(Q2)} a consistent difference between human and LLM post stance distributions. With the exception of Gemini-3-Flash-Preview and GPT-5-Mini (for (\textit{strongly disagree}), LLMs overestimate the neutral position in the simulated post-stances and underestimate the frequency of extreme views in humans (\textit{strongly disagree} or \textit{strongly agree}). LLMs appear to regress to a more neutral position overall.

\begin{figure*}[ht]
  \centering
  \includegraphics[width=\linewidth]{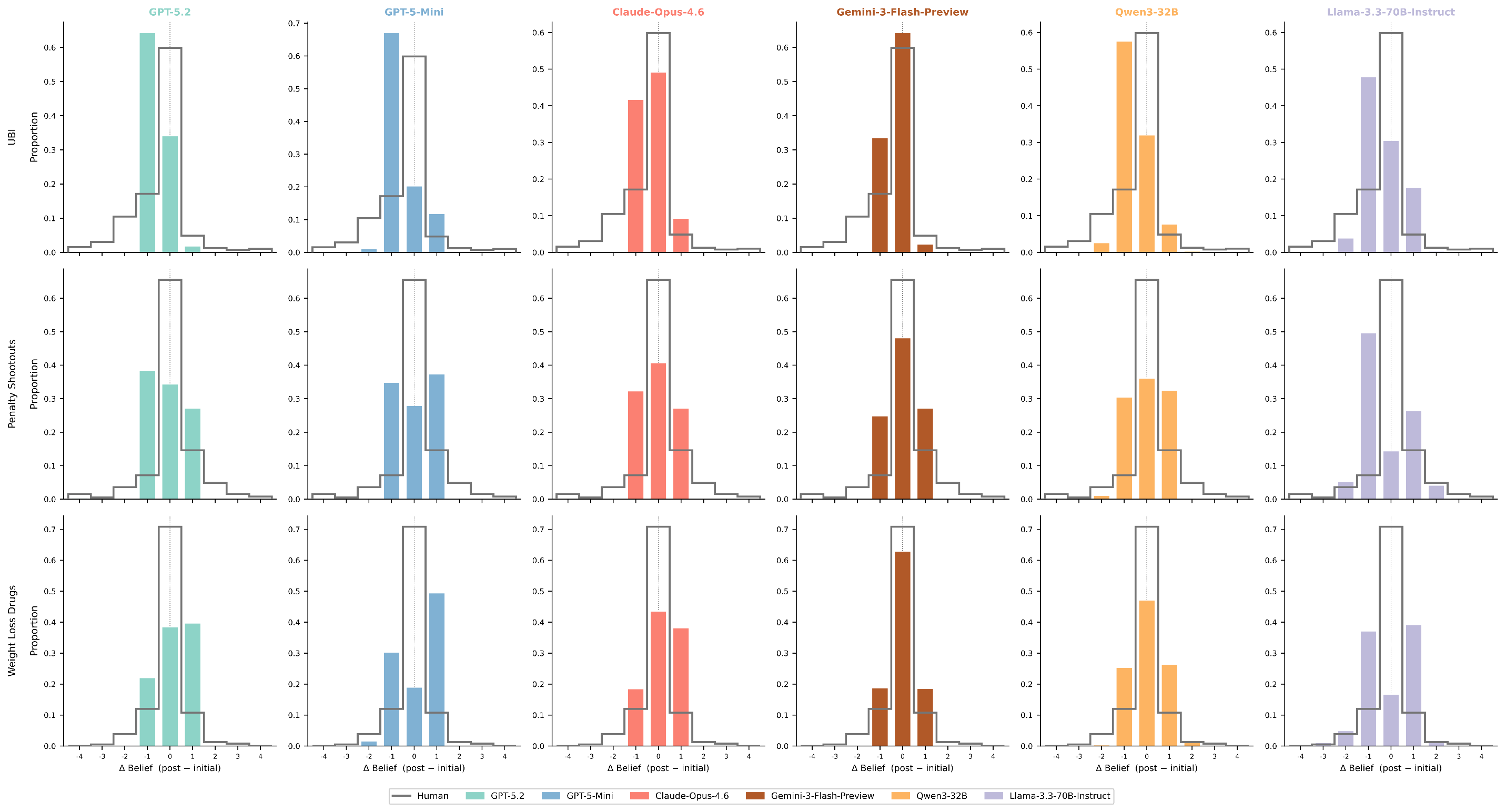}
  \caption{\textbf{Belief Change Distribution by Topic ($\Delta$ = post-stance - initial belief).} Across every topic, humans are far more likely than LLMs to leave their stance unchanged ($\Delta=0$), while LLMs spread more mass onto small nonzero deltas ($\Delta=\pm 1$). This underpins the finding that LLMs produce belief updates more frequently than humans, who more often retain their initial stance.}
  \label{fig:belief_delta_by_topic}
\end{figure*}

Second, we ask whether LLMs reproduce the variability in belief updates observed among human participants. As can be seen in Figure~\ref{fig:results_1_overview}i, the belief updates (\textit{post stance - initial stance}) simulated by LLMs have a slightly larger mean absolute value overall, but a smaller standard deviation than the human ground truth data. Two-sided Brown-Forsythe tests show that the standard deviation of human absolute belief change (SD = 0.87) differs significantly from than that of all six LLMs (all $p < 0.0002$, except Claude-Opus-4.6: $p = 0.0047$; all below the Bonferroni-corrected threshold $\alpha_\text{corrected} = 0.05/6 \approx 0.0083$). LLMs produced belief updates more frequently than humans, who had a tendency to keep their initial stance more often after reading the comments. But when humans did update their beliefs, they tended to have greater differences between their initial and their post-stance than LLMs. This can also be observed in Figure~\ref{fig:belief_delta_by_topic}, which plots the histogram of the stance changes of LLMs against the human reference data split by the three discussion topics.

Third, we compared how humans and LLMs rated the convincingness of individual comments. We computed Kendall rank correlation coefficients between human and LLM rankings. Details on these coefficients are given in the methods section. As can be seen in Table~\ref{tab:kendall rank} the correlation coefficients for all models are close to zero, indicating that the LLMs' rankings of the comments are more or less independent of the human ground truth rankings.

\begin{table}[t] \centering
\caption{The correlation coefficient for the rankings of comments between LLMs and humans. The values indicate independence between LLM and Humans in ranking the comments in terms of convincingness.}\label{tab:kendall rank}
\begin{tabular}{c c}
\toprule
\textbf{Model} & \textbf{Mean Kendall Rank Correlation Coefficient}\\
\midrule
GPT-5.2 & -0.0190\\
GPT-5-Mini & -0.0020 \\
Gemini-3-Flash-Preview & -0.0023 \\
Claude-Opus-4.6 & 0.0196 \\
Qwen-3-32B & -0.0225\\
Llama-3.3-70B-Instruct & -0.0361\\
\bottomrule
\end{tabular}
\end{table}

While it is an open question whether LLMs would be swayed more by comments they rate as more convincing, this gives at least a tentative insight into whether LLMs simulate not only the results of belief updates, but can also indicate reasons humans might have for their belief updates.
However, we observe that all LLMs ranked the three comments they had read by how convincing they are independently of how humans had ranked them. 
\textit{Human-LLM} pairs do not agree on how to rank these comments. This difference could explain why LLMs struggle with simulating human belief updates.

To further compare the extent to which responses are internally consistent among humans and across LLM persona conditions, we computed the mean rank given to each comment, whenever it was ranked. There are three topics with three packages of three comments each, equaling 27 comments in total. Figure~\ref{fig:results_1_overview}b shows the average and standard deviation of the mean ranks for these 27 comments. A comment with a low mean rank was consistently rated as more convincing than other comments by the respective LLM or human raters. We observe in particular that the standard deviation is much smaller for humans than for the six LLMs. Comparing the spread of each model's mean-rank distribution against the human distribution using two-sided Brown-Forsythe tests, confirms that the variability of human rankings differs significantly from all six LLM models, with human rankings being less variable ($p < 0.0003$ for all LLMs). LLMs more consistently rated the same comments as more convincing or less convincing across the personas they were conditioned on, whereas humans tended to disagree with each other more on which comments are more or less convincing.

\subsection*{Individual predictors of belief updates}
In the previous section, we found that human and LLMs rankings of individual comments are largely independent, suggesting that humans and LLMs are not responsive to comment-level convincingness in the same way. This raises the broader question of which input features drive belief updates in humans and LLMs and whether they are responsive to these features in similar ways.

\begin{figure}[t]
    \centering
    \includegraphics[width=\linewidth]{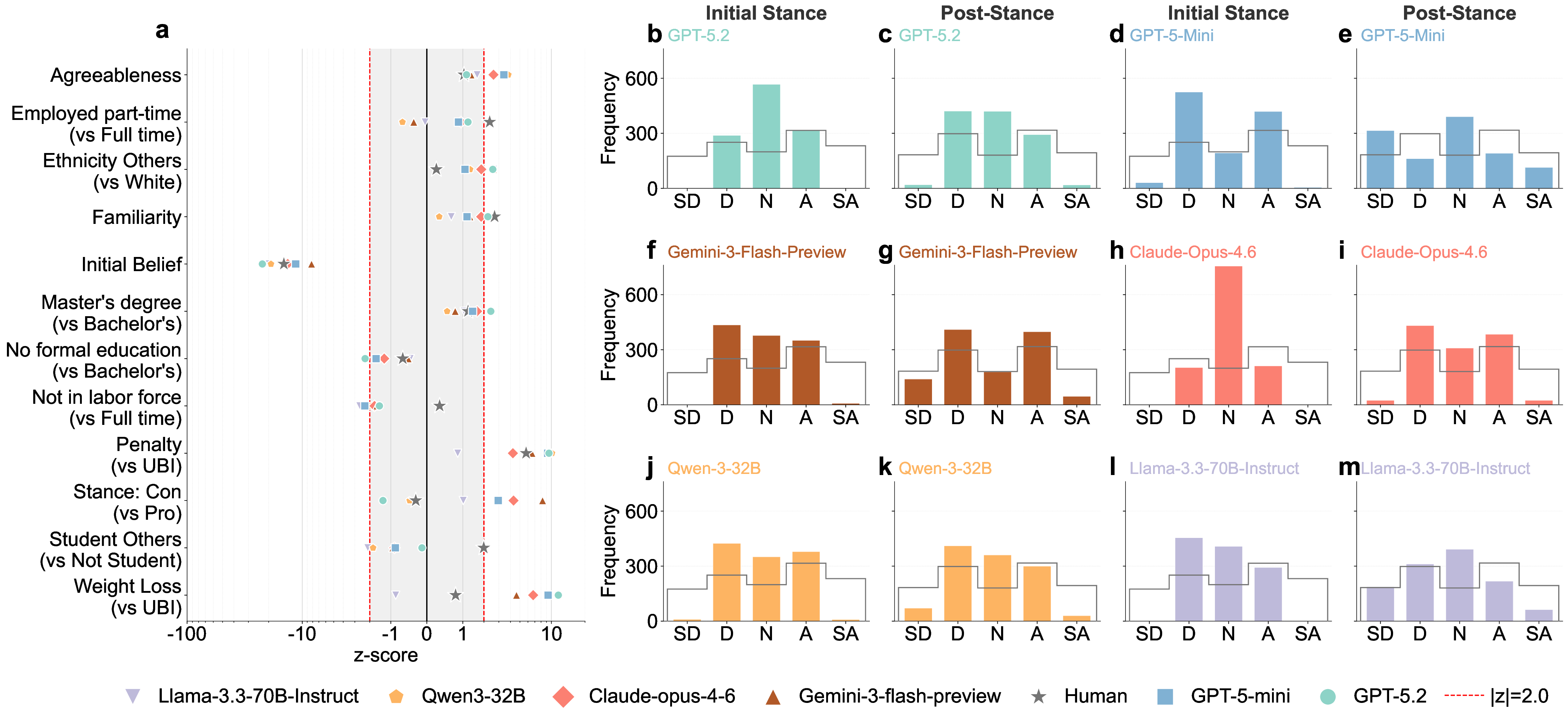}
    \caption{\textbf{Results of the mixed effects model and experiments based on simulated initial stances.} SD=strongly disagree, D=disagree, N=neutral, A=agree, SA=strongly agree. \textbf{a,} The panel shows the z-scores from the mixed effects model for all predictors where humans or at least one LLM exceeded the screening threshold of $|z| \geq 2$ (red dashed lines). Humans are highlighted by diamonds. All six LLMs are shown by colored circles. A z-score beyond the threshold indicates that this dimension as a measurable effect on belief updates. \textbf{b–m,} These panels show additional experiments in which LLMs first simulated the initial stances provided all other relevant individual information and then updated them after reading the same comments as human participants. LLMs fail to produce human-like initial-stance distributions, and under this condition they subsequently also all fail to reproduce the human post-stance distributions. This contrasts with our earlier findings, where some LLMs matched the human post-stance distributions closely when given the true initial stances. The gray outlines in each plot show the human distributions for reference.}
    \label{fig:results_2_features}
\end{figure}

To explore this question we used mixed effects models, testing which demographic, personality, and contextual features predicted belief updates in humans and in LLM simulations. To identify features with potentially meaningful effects, we highlight predictors for which humans or any LLM showed a z-score of $|z| \geq 2$. We use this threshold as a screening criterion rather than a formal significance test. The results are shown in Figure~\ref{fig:results_2_features}a. By this criterion, demographic and personality trait attributes showed little effect on human belief change. The features that stood out were initial stance, topic, topic familiarity, and employment status (employed part-time vs.\ employed full-time). That the initial stance has a large effect is expected logically (from an initial stance of 2 only differences $\leq0$ to the post-stance are possible) and based on the fact that participants only read three social media comments in our study compared to all their related background knowledge. Notably, apart from one employment status condition, no demographic or personality trait attribute showed a meaningful effect. If these attributes do not drive human belief updates in our setting, conditioning LLMs on them is unlikely to improve simulation fidelity.


Some LLMs showed sensitivity to predictors that had no meaningful effect in the human sample. The personality trait \textit{Agreeableness} predicted belief changes for Claude-Opus-4.6 and Qwen3-32B, \textit{ethnicity} and \textit{highest qualification} became predictors for GPT-5.2, \textit{topic phrasing} (positive vs.\ negative framing) influenced both Claude-Opus-4.6 and Gemini-3-Flash-Preview, and \textit{student status} slightly affected belief updates in Llama-3.3-70B-Instruct.

We looked into the effect of the topic framing in more detail. In humans, framing affected initial beliefs for \textit{Penalty Shootout} ($p=0.0043$) and \textit{Weight Loss Drugs} ($p=0.0077$; two-sided Mann--Whitney U tests, significant after Bonferroni correction), with pro-framed participants reporting more support for the topic. Since simulated participants start from the corresponding human's initial belief, these effects are shared by humans and LLMs by design. We therefore tested whether framing additionally affects belief updates, comparing the belief change between framing groups with two-sided Mann–Whitney U tests for each source and topic (Supplementary Table 4). Human updates show no significant framing effect for any topic. Among the LLMs, Gemini-3-Flash-Preview shows a significant framing effect for \textit{UBI} and \textit{Weight Loss Drugs}, Claude-Opus-4.6 for \textit{UBI}, and Llama-3.3-70B-Instruct for \textit{Weight Loss Drugs}. This is consistent with the mixed effects model, where Gemini-3-Flash-Preview and Claude-Opus-4.6 also showed the strongest stance effect. All significant effects are negative, indicating that, relative to the other framing, simulated beliefs shifted away from the statement wording participants were shown. For example, for UBI, Gemini-3-Flash-Preview agents shown the thesis that everyone should receive a UBI lowered their support by 0.54 points on average (from 0.39 to -0.15), whereas agents shown the opposing thesis barely changed (0.61 to 0.56; Claude-Opus-4.6: -0.50 vs. -0.13). Human participants shifted by similar amounts under both framings (-0.43 vs. -0.36).


\subsection*{The role of demographic and personality trait data}
These findings motivated our third research question \textbf{(Q3)} and lead us to test how much the persona's demographic and personality trait content contributes to LLM's simulation fidelity.

Demographic and personality trait data did not show a great effect on human or LLM belief updates in our mixed effects model. This raises the question how important for simulation fidelity it is to condition LLMs on personas based on these attributes. To investigate this, we designed additional experiments with three reduced persona templates that omitted demographics, personality traits, or both data and ran them on \textit{Qwen3-32B}, \textit{Llama-3.3-70B-Instruct}, and \textit{GPT-5-Mini} due to resource constraints. All reduced templates still contained participants' initial beliefs and topic familiarity. Table~\ref{tab:ablation_persona} shows the results of the permutation test on paired stance differences between human and LLM post stances for each condition.

For Llama-3.3-70B-Instruct the test consistently shows that the simulated post stances differ significantly from the human data. For Qwen3-32B and GPT-5 mini we did not reject the null hypothesis in the \textit{complete persona} setting, indicating that the models faithfully simulated the human sample on this measure. Only when we leave out demographic information from the persona for Qwen3-32B, does the simulation fidelity decrease to the extent that the p-value drops below the corrected threshold ($\alpha_{corrected}\approx0.0083$). However, leaving out even more information, demographic and personality trait data, increases simulation fidelity. In this setting, the simulated post stances did not differ significantly from the human post stances for Qwen3-32B, revealing some internal inconsistency in these results.
In fact, in all other cases, the permutation test still does not show a significant difference between simulated and human post stances even if the LLMs lack access to the demographic and personality trait data of the participant they are meant to simulate. 

Yet, while mostly below the level of significance, looking at the observed mean post stance differences we note that in Qwen3-32B leaving out persona information consistently increases the observed difference, indicating a greater dissimilarity of the simulated belief updates to the human ground truth data. In GPT-5-Mini leaving out persona information has no consistent impact on the observed difference. We again observe some internal inconsistency. Leaving out both demographic and personality trait data from the persona decreases the absolute observed difference, indicating that the simulation became more accurate without this information. While leaving out only personality data increases it, indicating an improvement of the simulation quality.

Overall, we find that leaving out demographic and personality trait information from the persona has no consistent impact on simulation fidelity. According to the permutation test on paired stance differences, Qwen3-32B and GPT-5 mini achieve their maximal simulation quality already when they only have access to the participants initial stance, topic familiarity and the comments they read, i.e. without access to their demographic and personality trait data.

\begin{table}[ht] \centering
\caption{Permutation test on paired stance differences results for LLM post-stances based on different persona information conditions against human post-stances. Each cell shows the observed difference with the p-value in parentheses. Lower observed differences indicate more similarity to human stances.}\label{tab:ablation_persona}
\begin{tabular}{l c c c}
\toprule
\textbf{Persona information} & \textbf{Qwen3-32B} & \textbf{Llama-3.3-70B-Instruct} & \textbf{GPT-5 mini} \\
\midrule
Complete persona        & 0.055 (0.081) & 0.118 (0.001) & 0.011 (0.753) \\
Without demographic     & 0.083 (0.008) & 0.112 (0.001) & -0.003 (0.959) \\
Without personality     & 0.060 (0.061) & 0.117 (0.001) & 0.055 (0.104) \\
W.o. dem. and pers.     & 0.073 (0.023) & 0.129 ($<$0.001) & -0.005 (0.898) \\
\bottomrule
\end{tabular}%
\end{table}


\subsection*{The need for accurate initial stances}
While demographic and personality trait information is often assumed to improve persona based simulation fidelity \cite{Argyle_Busby_Fulda_Gubler_Rytting_Wingate_2023, jiang2023evaluating, jiang-etal-2024-personallm}, our results provide no evidence that including these attributes yields a systematic or robust benefit for aligning LLM generated post stances with human post stances. Initial stances, on the other hand, were as expected very indicative of human belief updates. To further investigate this, we ran additional experiments testing the impact of initial stances on simulation fidelity. 

First, we tested whether LLMs were able to predict initial beliefs based on full persona information, including each participants topic familiarity, demographic, and personality trait data. The initial stance panels in Figure~\ref{fig:results_2_features}~b-m (Initial Stance) show the distributions of these simulated initial beliefs against the outline of the human initial stance distribution. We find that LLMs did not accurately simulate the distribution of initial beliefs. Chi-squared tests of independence confirm that the simulated initial stance distributions differed significantly from the human distribution for all six models (all $p < 0.0001$). As we had already observed on the simulated post stances, they underestimate the prevalence of more extreme positions (strongly agree or strongly disagree).

Second, we tested the extent to which LLMs were still able to simulate human post stances, when they had to base their simulation not on the participants actual initial stances, but on the initial stances they had simulated for each participant using their topic familiarity, demographic, and personality trait data. The results of these simulations can be seen in the post stance panels in Figure~\ref{fig:results_2_features}~b-m. Running the permutation test on paired stance differences again in this setting, we get a p-value of $p<0.0001$ for all models. When LLMs have to simulate both the initial and the post stance, we can reject the null hypothesis that LLMs produce faithful belief updates corresponding to the human ground truth data.

\section*{Discussion}

The use of LLMs as a substitute for human study participants has gained considerable momentum across the behavioral and social sciences. It promises to reduce the cost and time of empirical research while enabling interventions on systems that would otherwise be difficult to study directly \cite{park2023generative, YaxAnlloPalminteri2024Reasoning, gao2024large, Wang2024asurvey, xie2024can, cui2025largescale, Sakamoto2025value, wang-etal-2025-decoding, tang-etal-2025-gensim, piao2025agentsocietylargescalesimulationllmdriven, GrizzardEtAl2025ChatGPTMoralJudgments, zeng2026toohuman}. Social media debate simulations represent one prominent instantiation of this trend, deploying LLM agents to study polarization dynamics and evaluate intervention strategies at a scale traditional human studies cannot match \cite{chuang-etal-2024-simulating, gao2024large, Wang2024asurvey, wang-etal-2025-decoding}. Yet the validity of conclusions drawn from any such simulation depends on whether the individual behavioral mechanisms being modeled actually resemble human behavior. Prior studies have largely sidestepped this question. They validate simulations only at the aggregate level against outcome distributions or mathematical models rather than testing whether individual simulation steps faithfully reproduces what a human would do \cite{wang-etal-2025-decoding, chuang-etal-2024-simulating, zhang-etal-2025-ga}. We addressed this gap for one key step, isolating belief updating in social media debate simulations, where an agent reads peer messages and revises its stance on some discussion topic. We investigated whether LLMs can replicate this process for matched human participants under controlled conditions.

\textbf{Empirical Grounding.} All tested LLMs failed to simulate participants' initial stances, and without access to the actual ones, they produced belief updates that differed significantly from those of humans. Providing the actual initial stances removed the detectable systematic bias for some LLMs. Even then, further evidence suggests that the details of the update process differ: LLMs tend to update their beliefs more often, but also more incrementally, and they fail to predict which comments humans rank as most convincing.

Our findings have two implications. First, they imply that caution is needed in current simulation setups of social media debate dynamics, which span multiple communication rounds \cite{chuang-etal-2024-simulating, gao2024large, wang-etal-2025-decoding, piao2025emergencehumanlikepolarizationlarge, piao2025agentsocietylargescalesimulationllmdriven}. In our experiments, LLM simulations showed no statistically significant deviation from human belief updates only when they were conditioned on realistic initial stances. As the simulations continue, errors may compound, a phenomenon we term \textit{simulation drift}. For later communication rounds, there is no guarantee that the initial stances in that round are still human-like even if the simulation was initialized with actual humans' stances. Our experiments suggest that as initial stances move away from a human starting point, the fidelity of the simulation becomes continuously worse. This problem is compounded by the common practice of initializing agents with random rather than empirically grounded initial beliefs \cite{chuang-etal-2024-simulating, wang-etal-2025-decoding}, meaning simulation errors are present from the very first round.

Second, our results have implications for the wider field of LLM-based simulations of human study participants. Our experiments show that simulations lacking grounding in realistic, context-relevant data diverge from human behavior, suggesting that such grounding is a precondition for any attempt at faithful simulation. In our study of belief updates that meant grounding the topic familiarity ratings and realistic initial stances. It is common practice to condition LLM agents on \textit{personas} that individuate the agents and diversify their behavior. These personas are generally based on key demographic attributes such as \textit{age}, \textit{gender}, \textit{ethnicity}, \textit{education} and \textit{occupation} \cite{chuang-etal-2024-simulating, xie2024can} and sometimes also include the Big Five personality traits \cite{goldberg1990bigfive, wang-etal-2025-decoding}. In studies on LLM-based user simulations for recommender systems personas additionally include domain specific information such as movie or genre preferences or watch histories \cite{Zhu-2024-how, wang2025user}. While LLMs have previously been shown to reflect personality trait data \cite{SerapioGarcia2025PsychometricLLM, Sakamoto2025value, sakai2026effects}, demographic and personality trait attributes were not a significant contributor to simulation fidelity in our experiments. This corroborates earlier results, where additional demographic data failed to help LLMs match human reasoning depth in game theoretic settings \cite{gao2025takecaution}. Our experiments indicate that researchers should not rely on demographic and personality trait attributes by default, but go beyond them to identify which features are critical to simulation accuracy in their specific setting (initial stances in our case). They should ensure that their simulation is grounded in realistic distributions along these features.

A similar observation has been made by Sakamoto et al. \cite{Sakamoto2025value}, who studied trust behavior in LLM agents conditioned on values of varying abstraction, across language (English vs. Japanese) and dialogue topic (hobbies vs. housing). They found that the culturally more nuanced factor of language only produced a significant effect when agents were conditioned on more specific, lower-level values. When conditioned on broader, abstract higher-order values, the language effect disappeared.

\textbf{Model Suitability.} That we found significant differences between simulated and human post stance pairs for some, but not all LLMs also raises the question of what makes a model better at simulating human study participants. Our main experiments cover several model families and sizes. To probe this further, we ran an additional sweep over the Qwen3.5 and Gemma-4 families, spanning 0.8B to 31B active parameters (up to 122B total for the mixture-of-experts models). Supplementary Table~16 reports, for each model, the mean paired post-stance difference to the human participants with its permutation test, and the mean absolute error (MAE), by which the models are ranked.

Whether a model is systematically biased shows no consistent relation to its size. Within Qwen3.5, the 2B, 4B, and 122B-A10B models differ significantly from the human participants, while the 0.8B, 9B, 35B-A3B, and 27B models do not. The MAE, in contrast, generally decreases with scale. The largest dense model in each family has that family's lowest MAE (Gemma-4 31B: 0.69; Qwen3.5 27B: 0.78), and within Gemma-4 the MAE falls monotonically with active parameters. The trend is not strict: Qwen3.5-2B is less accurate than Qwen3.5-0.8B, and both Qwen3.5 mixture-of-experts models trail the dense 27B model despite their larger total size, performing in line with their active parameter count instead.

Further, we observed a general bias in LLMs to favor more neutral positions. This may be caused by model post training, where models are trained to follow human preferences, which might bias them to be more agreeable. Prior work similarly asks whether reinforcement learning from human feedback makes LLMs more cooperative in strategic games \cite{pal2026large}. We tested whether post-training impacts simulation fidelity in the OLMo 3 model family \cite{olmo2025olmo3}, for which post-training checkpoints are available. However, we did not find a general trend, where models became more or less accurate simulators during post-training. Identifying what makes a model a better simulator of human study participants is a promising direction for future work.

\textbf{LLM Biases.} Beyond our core findings on simulation fidelity and their methodological implications, our experiments revealed several systematic differences between human and simulated belief updates. Some of which align with prior findings, while others diverge, contributing to a wider understanding of how LLMs model human participants in experimental settings.

Our experiments showed a clear bias in LLMs towards neutral positions, underestimating the frequency of more extreme views, in particular when tasked with simulating human initial stances. This supports the broader observation that LLMs do not reflect the variability of human data, though it stands in contrast to findings of increased extremity in LLM outputs. Acrebi et al. \cite{Acrebi2023large} found that LLMs exhibit social, negative, and threat-related biases in ways that parallel human tendencies. Wang et al. \cite{wang-etal-2025-decoding} observed slightly higher polarization levels in LLM-based simulations compared to classical numerical models. And Grizzard et al. \cite{GrizzardEtAl2025ChatGPTMoralJudgments} found that LLMs make more extreme moral judgments than humans while clustering around a restricted number of values, failing to reflect human variability. The disagreement between our findings and prior indications of increased polarization calls for a closer investigation into the conditions under which LLMs tend toward neutrality versus extremity.

We further found that while humans had a greater standard deviation in the magnitude of their belief shifts, LLMs produced belief updates more frequently than humans. This is consistent with a bias in instruction-tuned LLMs to produce an action when the prompt mentions the option to do so, though we did not directly test this mechanism. A similar pattern was observed by Cui et al. \cite{cui2025largescale}, who found that LLMs faithfully replicated main and interaction effects in replications of psychology and management experiments, but consistently showed larger effect sizes than human participants. Rieger and Roesler \cite{rieger2026ChatGPT} similarly found that LLM agents in an advice-taking task adjusted their initial estimates towards the advice they received far more readily than human participants and showed markedly lower variability, rarely retaining their initial judgment.

Our findings underscore that simulation fidelity cannot be assumed. LLMs showed systematic biases toward neutrality, updated beliefs more readily than humans, and required empirically grounded initial stances to produce human like updates at all. In particular, for simulations in which agents continually update their beliefs, our experiments suggest that simulation drift represents a severe risk. To minimize that risk, researchers need to ensure the fidelity of their simulation through careful experimental design, empirical grounding, and rigorous validation at the level of individual simulation steps.

\section*{Methods}
\subsection*{Human Data Collection}
\begin{table}[h]
\centering
\caption{Demographic characteristics of the 391 study participants.}
\label{tab:demographics}
\begin{tabular}{lrr}
\toprule
\textbf{Characteristic} & \textbf{$n$} & \textbf{\%} \\
\midrule
\textit{Gender} & & \\
\quad Female      & 202 & 51.7 \\
\quad Male        & 188 & 48.1 \\
\quad Diverse     &   1 &  0.3 \\
\midrule
\textit{Age} & & \\
\quad 18--24      &  38 &  9.7 \\
\quad 25--34      &  69 & 17.6 \\
\quad 35--44      &  67 & 17.1 \\
\quad 45--54      &  68 & 17.4 \\
\quad 55--64      &  98 & 25.1 \\
\quad 65+         &  51 & 13.0 \\
\midrule
\textit{Ethnicity} & & \\
\quad White                          & 332 & 84.9 \\
\quad Asian                          &  28 &  7.2 \\
\quad Black / African descent        &  14 &  3.6 \\
\quad Multiple ethnic groups         &   8 &  2.0 \\
\quad Other ethnic group             &   5 &  1.3 \\
\quad Hispanic or Latino             &   2 &  0.5 \\
\quad Middle Eastern / North African &   2 &  0.5 \\
\bottomrule
\end{tabular}
\end{table}

We collected our human ground truth data via Prolific. We recruited 430 participants from the UK in total, of whom 391 completed all three topics of the survey on time. Of the remaining 39 participants, 7 dropped out at the consent form, an additional 26 dropped out during the first topic, 5 dropped out during later topics and 1 participant completed the survey only after we had set a data collection cutoff point. In the following, we only report results for the 391 participants who completed the survey on time. These participants were, as guaranteed by Prolific, representative of the total UK population in terms of gender, age, and ethnicity. Table~\ref{tab:demographics} shows the distributions across these dimensions. The order in which the three topics were presented was randomized for each participant. For each topic, the participant was first presented with a statement (negative variants in bold):

\begin{enumerate}
    \item \textit{"Everyone should receive Universal Basic Income."} \textit{\textbf{["There should be no Universal Basic Income."]}}
    \item \textit{"Penalty shootouts are a good \textbf{[bad]} way to determine the winners of football matches."}
    \item \textit{"Using weight loss drugs like Ozempic is a good \textbf{[bad]} way to lose weight."}
\end{enumerate}

Each participant was either presented with the statement in support of or opposition to the topic at random. Afterwards they were asked to rate their initial stance on the statement on a 5-point Likert scale from $(-2)$ \textit{strongly disagree} to $(2)$ \textit{strongly agree} and how familiar they are with the topic. We selected three topics that are as distinct from one another as possible and on which opinions vary. We further aimed for topics that participants would find familiar and engaging, but that are not so closely tied to personal identity as to preclude opinion change. Initially, participants were only told that we were presenting them with curated messages packages. Only after data collection we informed them that they had been presented with comments sourced from Reddit that may have contained some false information. To keep the setting comparable, LLMs were also only instructed that the given comments had been presented to the participant they are tasked to simulate, not giving away the data source.

Next, participants were given a package of three \textit{r/changemyview} messages on the discussion topic. We manually curated these messages, since many posts on the subreddit are only intelligible in the context of the thread they respond to. We selected messages that were understandable on their own, varied in stance, and were of sufficient quality to plausibly change participants' minds. For each topic, we assembled three packages: one with two messages supporting the \textit{(positively phrased)} claim and one opposing it, one with two opposing and one supporting message, and one consisting of more neutral messages. Each participant was randomly assigned one package. 

After having read the messages we asked participants to rate their post-stance on the topic on the same 5-point Likert scale, to rank the three messages they had read by how convincing they found them, to judge the general public's stance on the discussion topic and to give a reason they had for updating or sticking to their initial belief. A brief analysis of the general public's stance judgments and reasons given for belief updates by humans and LLMs is given in the Supplementary Information. In their simulation of higher order beliefs LLMs exhibited the same previously reported trend of overestimating neutral positions. The reasons for belief updates they gave tended to be longer and more detailed than the reasons given by humans. We assume that this is the result of LLMs being trained to diligently comply with the instructions given to them. Beyond this qualitative observation we did not distill general insights from the provided reasons and focused on reporting the quantitative results. 

This data collection procedure was repeated for three topics, such that each participant was presented with each topic, albeit in random order. Finally, we recorded participants' demographic data such as gender, age, occupation and asked them to fill out a 10-item reduced Big Five personality test \cite{rammstedt2007bfi10}.

The resulting dataset contains 1173 human belief update steps (391 participants $\times$ 3 topics), spanning 27 messages in total (3 topics $\times$ 3 packages $\times$ 3 messages per package); each participant was presented with 3 messages per topic, i.e. 9 messages in total. Each data instance captures one participant–topic pair: the statement variant and message package the participant was assigned, their initial stance and topic familiarity, their post-stance and comment rankings, and their demographic and personality trait profile.

Prior to data collection this project has been reviewed by the research ethics committee of the Interdisciplinary Transformation University Austria under the case number 2025-09. The experiment pre-registration is available at: https://doi.org/10.17605/OSF.IO/QJEMZ.


\subsection*{LLM Data Collection}
For our simulation experiments, we used six LLMs of varying sizes. GPT-5.2 and GPT-5-Mini \cite{singh2025openaigpt5card}, Claude-Opus-4.6 \cite{anthropic2026claudeopus46}, Gemini-3-Flash-Preview \cite{google2025gemini3flash}, Qwen3-32B \cite{yang2025qwen3technicalreport} and Llama-3.3-70B-Instruct \cite{grattafiori2024llama3herdmodels}. OLMo3 \cite{olmo2025olmo3} Qwen3.5 \cite{qwen3.5} and Gemma-4 \cite{gemmateam2026gemma4} models were used for additional experiments on the effect of post-training and model size.

For each human study participant and topic we prepared an input that combined all the information the human study participant had been presented with, when updating their belief on the given topic, i.e. we ran our simulation on a participant by participant basis. The LLM always received the topic with the same pro or con formulation the current human participant had seen. The prompt context contained the initial stance and topic familiarity the human participant had recorded for the given topic, the LLM received the same comments the participant had seen and additionally we constructed a persona for each human participant based on their demographic and personality data. Given this context the LLMs were instructed to simulate the human participants belief update and to output the post-stance, the ranking of the three presented comments, a rating of what the simulated participant would judge the general public's stance to be and a reason for belief change in the voice of the participant in a structured json format. The prompt template we used can be seen in Supplementary Figure~9.

By building our personas 1-to-1 from human participants' data we ensure that our LLM simulations are representative of the sample they are intended to simulate. As a result, we record paired data, where each human belief update is linked to a direct simulation for each of the LLMs we used.

For our experiments we used the standard decoding settings for each model's API and a temperature of 0.7. We tested the impact of changing the model temperature to 0.0 or 2.0 on \textit{GPT-5-Mini,} \textit{Qwen3-32B} and \textit{Llama-3.3-70B-Instruct}. The results are presented in the Supplementary Information. With higher temperatures the generation quality suffers in \textit{Qwen3-32B} and \textit{Llama-3.3-70B-Instruct}, leading to some non-sense generations and output format violations. But we found no clear impact on whether the simulations were systematically biased. This may partly be due to the fact that we instruct models to give their outputs in a structured json format.

The comments used in our experiments are publicly available and may therefore be part of the training data of the models we evaluate. For the OLMo models used in our post-training ablations, whose pre-training corpora are publicly released, this could in principle be verified directly; for the remaining models, training data is not disclosed. Such contamination may extend beyond the comments themselves to users' reactions to them, which would plausibly make models more human-like in their simulated belief updates. Contamination should therefore bias our results towards agreement with human behavior, not away from it. The significant differences between LLM and human belief updates we frequently observe are thus, if anything, conservative. Conversely, where we do not detect significant differences, we cannot rule out that this reflects memorization rather than a genuine capacity for simulating human belief updates. This concern is supported by Gao et al.~\cite{gao2025takecaution}, who find that LLM agents behave in a more human-like manner in well-known game-theoretic games than in less familiar ones. This suggests that human-likeness may partly stem from memorization and may not generalize to out-of-distribution tasks.

\subsection*{Paired stance differences}
To directly compare human to simulated post-stances we used a two sided permutation test on the mean of paired differences between the LLM simulation and the corresponding human post-stance as the test-statistic. Our null hypothesis (\(H_0\)) was that LLMs produced faithful belief updates corresponding to human ground truth data. The alternative hypothesis (\(H_1\)) was that LLMs simulate belief updates that differ from the belief updates recorded from humans.

We chose the permutation test because it is a non-parametric test and only assumes the exchangeability of human and LLM post stances, or a flip within the pair under the null hypothesis. This would lead to a sign-flip in the  paired difference. We approximated the distribution of the mean difference under the null hypothesis using a Monte Carlo method, by repeatedly sampling sequences of paired differences and computing their means. The resulting $p$-value gives the probability of the observed mean difference being at least as extreme as the one in our data under the null hypothesis. The permutation test was carried out independently for all six LLMs.

There are significant differences between the paired test and distribution level tests such as the Chi-squared test. While the permutation test is more powerful in cases of directional shifts in means of distributions and when pair level performance is the goal of the model, the Chi squared test is better equipped at detecting differences in overall shapes of our distributions and variances \cite{Holt_Sullivan_2023}.

Observed differences, the paired effect size (Cohen's $d_z$) and exact $p$-values for all six models are reported in Supplementary Table 10. The corresponding statistics and standardized effect sizes for the other analyses in this paper are given in Supplementary Tables 11-14.

We observed a difference in the variance of absolute belief updates between humans and LLMs, where LLMs tended to have more belief updates but of smaller magnitude. None of our comparisons between humans and LLMs assume variance homogeneity. The paired permutation tests reduce each human-LLM pair to a single paired difference and thus never pool or compare the variances of the two groups. The Chi-squared and Brown-Forsythe tests do not assume equal variances, but are sensitive to differences in variance (the Brown-Forsythe test directly tests for them). The mixed-effects models are fitted separately for humans and for each LLM, so each source has its own residual and random-effect variances, and no data are pooled across sources.

\subsection*{Correlation of comment rankings}\label{sec:comment_rankings}
In our experiments, each \textit{human-LLM} pair that shares the same demographic and personality traits had seen and ranked the same three comments for each of the three topics. From these rankings we calculate the Kendall rank correlation coefficient for all pairs as follows:
\begin{equation*}
\tau = \frac{N_c - N_d}{N},
\end{equation*}
where $N_c$ is the number of concordant pairs, $N_d$ the number of discordant pairs, and $N$ the total number of pairs. Concordant pairs are pairs of comments that both the LLM and human rank in the same order; discordant pairs are those ranked in opposite order. Given that we have three comments, the coefficient takes the possible values of 1, 0.33, -0.33 and -1 for each \textit{human-LLM} pair. A value of 1 implies a completely identical and -1 a completely opposite ranking. A value close to 0 indicates independence in the rankings within the pair. We calculated the mean coefficient across all pairs.

\subsection*{Mixed-effects models for feature importance}\label{sec:methods_mixed_effects_model}
We fitted separate linear mixed-effects models for humans and each of the six LLMs to identify which predictors are associated with belief shifts (post-stance - initial stance). Fitting one model per group allows direct comparison of predictor strengths across humans and LLMs.

The fixed effects included topic (Universal Basic Income, Penalty Shootouts, Weight Loss; reference: Universal Basic Income), stance direction (pro vs.\ con; reference: pro), initial belief, age, and topic familiarity (both mean-centred), as well as six categorical demographic variables (gender, ethnicity, country of birth, student status, employment status, and highest qualification) and the five Big Five personality trait scores (all mean-centred). For ethnicity, country of birth, and student status, the modal category accounted for more than 80\% of responses (White, Great Britain, and not currently a student, respectively). For any variable meeting this 80\% dominance threshold, all remaining categories were merged into a single 'other' group, yielding a binary contrast between the dominant category and everyone else. Figure~\ref{fig:results_2_features}a reports z-scores only for predictors where at least one LLM or humans had $|z| \geq 2$; all predictors listed above were included in every model.

The z-scores are derived by dividing each coefficient by its standard error. A threshold of $|z| \geq 2$ corresponds roughly to the conventional bar for a suggestive association under a standard normal approximation, and provides a familiar and interpretable screening criterion rather than a formal significance test. This analysis is descriptive: we use the models to characterize which factors are associated with belief updates in humans and in each LLM, not to test whether effects differ between LLMs and humans.

To account for the repeated-measures structure of the data (each participant responded to three topics), we included a random intercept per participant and a random variance component for topic. Models were estimated by maximum likelihood as implemented in \texttt{statsmodels} \cite{seabold2010statsmodels}:

\begin{equation*}
y_{ij} = \beta_0 + \mathbf{x}_{ij}^\top \boldsymbol{\beta} + u_i + v_j + \varepsilon_{ij}
\end{equation*}

where $y_{ij}$ is the belief shift for participant $i$ on topic $j$,
$\beta_0$ is the intercept,
$\mathbf{x}_{ij}$ is the vector of fixed-effect predictors,
$\boldsymbol{\beta}$ the corresponding coefficients,
$u_i \sim \mathcal{N}(0, \sigma_u^2)$ the participant random intercept,
$v_j \sim \mathcal{N}(0, \sigma_v^2)$ the topic variance component,
and $\varepsilon_{ij} \sim \mathcal{N}(0, \sigma^2)$ the residual.

{\small
\setlength{\bibsep}{0\baselineskip} 
\bibliographystyle{naturemag}
\bibliography{sn-bibliography}

@inproceedings{wang-etal-2025-decoding,
    title = "Decoding Echo Chambers: {LLM}-Powered Simulations Revealing Polarization in Social Networks",
    author = "Wang, Chenxi  and
      Liu, Zongfang  and
      Yang, Dequan  and
      Chen, Xiuying",
    editor = "Rambow, Owen  and
      Wanner, Leo  and
      Apidianaki, Marianna  and
      Al-Khalifa, Hend  and
      Eugenio, Barbara Di  and
      Schockaert, Steven",
    booktitle = "Proceedings of the 31st International Conference on Computational Linguistics",
    month = jan,
    year = "2025",
    address = "Abu Dhabi, UAE",
    publisher = "Association for Computational Linguistics",
    url = "https://aclanthology.org/2025.coling-main.264/",
    pages = "3913--3923"
}

@article{cui2025largescale,
  title   = {A Large-Scale Replication of Scenario-Based Experiments in Psychology and Management Using Large Language Models},
  author  = {Cui, Z. and Li, N. and Zhou, H.},
  journal = {Nature Computational Science},
  volume  = {5},
  pages   = {627--634},
  year    = {2025},
  doi     = {10.1038/s43588-025-00840-7}
}

@article{FriedkinJohnsen1990,
  author  = {Friedkin, Noah E. and Johnsen, Eugene C.},
  title   = {Social Influence and Opinions},
  journal = {Journal of Mathematical Sociology},
  year    = {1990},
  volume  = {15},
  number  = {3--4},
  pages   = {193--206},
  doi     = {10.1080/0022250X.1990.9990069}
}

@article{DeffuantEtAl2000MixingBeliefs,
  author  = {Deffuant, Guillaume and Neau, David and Amblard, Frederic and Weisbuch, G{\'e}rard},
  title   = {Mixing Beliefs Among Interacting Agents},
  journal = {Advances in Complex Systems},
  year    = {2000},
  volume  = {3},
  number  = {01n04},
  pages   = {87--98},
  doi     = {10.1142/S0219525900000078}
}

@inproceedings{chuang-etal-2024-simulating,
    title = "Simulating Opinion Dynamics with Networks of {LLM}-based Agents",
    author = "Chuang, Yun-Shiuan  and
      others",
    editor = "Duh, Kevin  and
      Gomez, Helena  and
      Bethard, Steven",
    booktitle = "Findings of the Association for Computational Linguistics: NAACL 2024",
    month = jun,
    year = "2024",
    address = "Mexico City, Mexico",
    publisher = "Association for Computational Linguistics",
    url = "https://aclanthology.org/2024.findings-naacl.211/",
    doi = "10.18653/v1/2024.findings-naacl.211",
    pages = "3326--3346"
}

@inproceedings{Zhu-2024-how,
author = {Zhu, Lixi and Huang, Xiaowen and Sang, Jitao},
title = {How Reliable is Your Simulator? Analysis on the Limitations of Current LLM-based User Simulators for Conversational Recommendation},
year = {2024},
isbn = {9798400701726},
publisher = {Association for Computing Machinery},
address = {New York, NY, USA},
url = {https://doi.org/10.1145/3589335.3651955},
doi = {10.1145/3589335.3651955},
booktitle = {Companion Proceedings of the ACM Web Conference 2024},
pages = {1726–1732},
numpages = {7},
location = {Singapore, Singapore},
series = {WWW '24}
}

@inproceedings{
yao2025taubench,
title={$\tau$-bench: A Benchmark for Tool-Agent-User Interaction in Real-World Domains},
author={Shunyu Yao and Noah Shinn and Pedram Razavi and Karthik R Narasimhan},
booktitle={The Thirteenth International Conference on Learning Representations},
year={2025},
url={https://openreview.net/forum?id=roNSXZpUDN}
}

@misc{barres2025tau2benchevaluatingconversationalagents,
      title={$\tau^2$-Bench: Evaluating Conversational Agents in a Dual-Control Environment}, 
      author={Victor Barres and Honghua Dong and Soham Ray and Xujie Si and Karthik Narasimhan},
      year={2025},
      eprint={2506.07982},
      archivePrefix={arXiv},
      primaryClass={cs.AI},
      url={https://arxiv.org/abs/2506.07982}, 
}

@misc{piao2025emergencehumanlikepolarizationlarge,
      title={Emergence of human-like polarization among large language model agents}, 
      author={Jinghua Piao and others},
      year={2025},
      eprint={2501.05171},
      archivePrefix={arXiv},
      primaryClass={cs.SI},
      url={https://arxiv.org/abs/2501.05171}, 
}

@misc{piao2025agentsocietylargescalesimulationllmdriven,
      title={AgentSociety: Large-Scale Simulation of LLM-Driven Generative Agents Advances Understanding of Human Behaviors and Society}, 
      author={Jinghua Piao and others},
      year={2025},
      eprint={2502.08691},
      archivePrefix={arXiv},
      primaryClass={cs.SI},
      url={https://arxiv.org/abs/2502.08691}, 
}

@article{GrizzardEtAl2025ChatGPTMoralJudgments,
  author  = {Grizzard, Matthew and others},
  title   = {ChatGPT does not replicate human moral judgments: the importance of examining metrics beyond correlation to assess agreement},
  journal = {Scientific Reports},
  year    = {2025},
  volume  = {15},
  pages   = {40965},
  doi     = {10.1038/s41598-025-24700-6},
}

@article{YaxAnlloPalminteri2024Reasoning,
  author  = {Yax, Nicolas and Anll{\'o}, Hern{\'a}n and Palminteri, Stefano},
  title   = {Studying and Improving Reasoning in Humans and Machines},
  journal = {Communications Psychology},
  year    = {2024},
  volume  = {2},
}

@article{gao2024large,
  author    = {Gao, Chen and others},
  title     = {Large language models empowered agent-based modeling and simulation: a survey and perspectives},
  journal   = {Humanities and Social Sciences Communications},
  volume    = {11},
  number    = {1259},
  year      = {2024},
  doi       = {10.1057/s41599-024-03611-3},
  url       = {https://doi.org/10.1057/s41599-024-03611-3},
  publisher = {Springer Nature}
}

@inproceedings{zhang-etal-2025-ga,
    title = "$\text{GA-S}^3$: Comprehensive Social Network Simulation with Group Agents",
    author = "Zhang, Yunyao  and
      others",
    editor = "Che, Wanxiang  and
      Nabende, Joyce  and
      Shutova, Ekaterina  and
      Pilehvar, Mohammad Taher",
    booktitle = "Findings of the Association for Computational Linguistics: ACL 2025",
    month = jul,
    year = "2025",
    address = "Vienna, Austria",
    publisher = "Association for Computational Linguistics",
    url = "https://aclanthology.org/2025.findings-acl.468/",
    doi = "10.18653/v1/2025.findings-acl.468",
    pages = "8950--8970",
    ISBN = "979-8-89176-256-5"
}

@article{wang2025user,
author = {Wang, Lei and others},
title = {User Behavior Simulation with Large Language Model-based Agents},
year = {2025},
issue_date = {March 2025},
publisher = {Association for Computing Machinery},
address = {New York, NY, USA},
volume = {43},
number = {2},
issn = {1046-8188},
url = {https://doi.org/10.1145/3708985},
doi = {10.1145/3708985},
journal = {ACM Trans. Inf. Syst.},
month = jan,
articleno = {55},
numpages = {37}
}

@inproceedings{tang-etal-2025-gensim,
    title = "{G}en{S}im: A General Social Simulation Platform with Large Language Model based Agents",
    author = "Tang, Jiakai  and
      others",
    editor = "Dziri, Nouha  and
      Ren, Sean (Xiang)  and
      Diao, Shizhe",
    booktitle = "Proceedings of the 2025 Conference of the Nations of the Americas Chapter of the Association for Computational Linguistics: Human Language Technologies (System Demonstrations)",
    month = apr,
    year = "2025",
    address = "Albuquerque, New Mexico",
    publisher = "Association for Computational Linguistics",
    url = "https://aclanthology.org/2025.naacl-demo.15/",
    doi = "10.18653/v1/2025.naacl-demo.15",
    pages = "143--150",
    ISBN = "979-8-89176-191-9"
}

@article{rammstedt2007bfi10,
  author  = {Rammstedt, Beatrice and John, Oliver P.},
  title   = {Measuring personality in one minute or less: A 10-item short version of the Big Five Inventory in English and German},
  journal = {Journal of Research in Personality},
  year    = {2007},
  volume  = {41},
  number  = {1},
  pages   = {203--212},
  publisher = {Elsevier Science},
  doi     = {10.1016/j.jrp.2006.02.001}
}

@misc{singh2025openaigpt5card,
      title={OpenAI GPT-5 System Card}, 
      author={Aaditya Singh and others},
      year={2025},
      eprint={2601.03267},
      archivePrefix={arXiv},
      primaryClass={cs.CL},
      url={https://arxiv.org/abs/2601.03267}, 
}

@misc{yang2025qwen3technicalreport,
      title={Qwen3 Technical Report}, 
      author={An Yang and others},
      year={2025},
      eprint={2505.09388},
      archivePrefix={arXiv},
      primaryClass={cs.CL},
      url={https://arxiv.org/abs/2505.09388}, 
}

@misc{anthropic2026claudeopus46,
  author       = {Anthropic},
  title        = {Claude Opus 4.6 System Card},
  year         = {2026},
  howpublished = {\url{https://www-cdn.anthropic.com/c788cbc0a3da9135112f97cdf6dcd06f2c16cee2.pdf}},
  note         = {Accessed: 2026-04-06}
}

@misc{google2025gemini3flash,
  author       = {Google DeepMind},
  title        = {Gemini 3 Flash Model Card},
  year         = {2025},
  howpublished = {\url{https://storage.googleapis.com/deepmind-media/Model-Cards/Gemini-3-Flash-Model-Card.pdf}},
  note         = {Accessed: 2026-04-06}
}

@misc{grattafiori2024llama3herdmodels,
      title={The Llama 3 Herd of Models}, 
      author={Aaron Grattafiori and others},
      year={2024},
      eprint={2407.21783},
      archivePrefix={arXiv},
      primaryClass={cs.AI},
      url={https://arxiv.org/abs/2407.21783}, 
}

@inproceedings{xie2024can,
author = {Xie, Chengxing and others},
title = {Can large language model agents simulate human trust behavior?},
year = {2024},
isbn = {9798331314385},
publisher = {Curran Associates Inc.},
address = {Red Hook, NY, USA},
booktitle = {Proceedings of the 38th International Conference on Neural Information Processing Systems},
articleno = {501},
numpages = {56},
location = {Vancouver, BC, Canada},
series = {NIPS '24}
}

@inproceedings{bougie2025simuser,
  title={Simuser: Simulating user behavior with large language models for recommender system evaluation},
  author={Bougie, Nicolas and Watanabe, Narimawa},
  booktitle={Proceedings of the 63rd Annual Meeting of the Association for Computational Linguistics (Volume 6: Industry Track)},
  pages={43--60},
  year={2025}
}

@article{gilbert2000agent,
  author  = {Gilbert, Nigel and Terna, Pietro},
  title   = {How to build and use agent-based models in social science},
  journal = {Mind \& Society},
  volume  = {1},
  pages   = {57--72},
  year    = {2000},
  doi     = {10.1007/BF02512229}
}

@article{goldberg1990bigfive,
  author  = {Goldberg, Lewis R.},
  title   = {An Alternative "Description of Personality": The Big-Five Factor Structure},
  journal = {Journal of Personality and Social Psychology},
  volume  = {59},
  number  = {6},
  pages   = {1216--1229},
  year    = {1990},
  doi     = {10.1037/0022-3514.59.6.1216}
}

@misc{olmo2025olmo3,
title={Olmo 3},
author={Allyson Ettinger and others},
year={2025},
eprint={2512.13961},
archivePrefix={arXiv},
primaryClass={cs.CL},
url={https://arxiv.org/abs/2512.13961},
}

@inproceedings{park2023generative,
author = {Park, Joon Sung and others},
title = {Generative Agents: Interactive Simulacra of Human Behavior},
year = {2023},
isbn = {9798400701320},
publisher = {Association for Computing Machinery},
address = {New York, NY, USA},
url = {https://doi.org/10.1145/3586183.3606763},
doi = {10.1145/3586183.3606763},
booktitle = {Proceedings of the 36th Annual ACM Symposium on User Interface Software and Technology},
articleno = {2},
numpages = {22},
location = {San Francisco, CA, USA},
series = {UIST '23}
}

@article{zeng2026toohuman,
  author    = {Yongchao Zeng and Calum Brown and Mark Rounsevell},
  title     = {Too human to model: the uncanny valley of large language models in simulating human systems},
  journal   = {npj Complexity},
  year      = {2026},
  volume    = {3},
  number    = {1},
  doi       = {10.1038/s44260-026-00075-1},
}

@article{SerapioGarcia2025PsychometricLLM,
  author    = {Gregory Serapio-Garc{\'i}a and others},
  title     = {A psychometric framework for evaluating and shaping personality traits in large language models},
  journal   = {Nature Machine Intelligence},
  year      = {2025},
  volume    = {7},
  number    = {12},
  pages     = {1954--1968},
  doi       = {10.1038/s42256-025-01115-6},
  url       = {https://doi.org/10.1038/s42256-025-01115-6}
}

@article{
Acrebi2023large,
author = {Alberto Acerbi  and Joseph M. Stubbersfield },
title = {Large language models show human-like content biases in transmission chain experiments},
journal = {Proceedings of the National Academy of Sciences},
volume = {120},
number = {44},
pages = {e2313790120},
year = {2023},
doi = {10.1073/pnas.2313790120},
URL = {https://www.pnas.org/doi/abs/10.1073/pnas.2313790120},
eprint = {https://www.pnas.org/doi/pdf/10.1073/pnas.2313790120},
}

@article{Sakamoto2025value,
  author  = {Sakamoto, Yuki and Uchida, Takahisa and Ishiguro, Hiroshi},
  title   = {Value-based large language model agent simulation for mutual evaluation of trust and interpersonal closeness},
  journal = {Scientific Reports},
  year    = {2025},
  volume  = {15},
  number  = {1},
  pages   = {41653},
  doi     = {10.1038/s41598-025-25531-1},
  url     = {https://doi.org/10.1038/s41598-025-25531-1},
  issn    = {2045-2322}
}

@article{Wang2024asurvey,
  author  = {Wang, Lei and others},
  title   = {A survey on large language model based autonomous agents},
  journal = {Frontiers of Computer Science},
  year    = {2024},
  volume  = {18},
  number  = {6},
  pages   = {186345},
  doi     = {10.1007/s11704-024-40231-1},
  url     = {https://doi.org/10.1007/s11704-024-40231-1},
  issn    = {2095-2236}
}

@article{yan2025social,
author = {Yan, Zihan and Xiang, Yaohong},
title = {Social Life Simulation for Non-Cognitive Skills Learning},
year = {2025},
issue_date = {May 2025},
publisher = {Association for Computing Machinery},
address = {New York, NY, USA},
volume = {9},
number = {2},
url = {https://doi.org/10.1145/3711068},
doi = {10.1145/3711068},
journal = {Proc. ACM Hum.-Comput. Interact.},
month = may,
articleno = {CSCW170},
numpages = {44}
}

@misc{reddit_cmv,
  author = {Reddit},
  title = {r/changemyview},
  howpublished = {\url{https://www.reddit.com/r/changemyview/}},
  note = {Accessed: 2026-01-12}
}

@inproceedings{seabold2010statsmodels,
  title={statsmodels: Econometric and statistical modeling with python},
  author={Seabold, Skipper and Perktold, Josef},
  booktitle={9th Python in Science Conference},
  year={2010},
}

@article{pal2026large,
    author = {Pal, Saptarshi and Mallela, Abhishek and Pracher, Lenz and Wei, Chiyu and Fu, Feng and Schnell, Santiago and Nowak, Martin A},
    title = {Large language models instantiate evolutionarily robust strategies of cooperation},
    journal = {PNAS Nexus},
    volume = {5},
    number = {6},
    pages = {pgag210},
    year = {2026},
    month = {06},
    issn = {2752-6542},
    doi = {10.1093/pnasnexus/pgag210},
    url = {https://doi.org/10.1093/pnasnexus/pgag210},
    eprint = {https://academic.oup.com/pnasnexus/article-pdf/5/6/pgag210/68510197/pgag210.pdf},
}

@misc{rilla2025recognisinganticipatingmitigatingllm,
      title={Recognising, Anticipating, and Mitigating LLM Pollution of Online Behavioural Research}, 
      author={Raluca Rilla and Tobias Werner and Hiromu Yakura and Iyad Rahwan and Anne-Marie Nussberger},
      year={2025},
      eprint={2508.01390},
      archivePrefix={arXiv},
      primaryClass={cs.CY},
      url={https://arxiv.org/abs/2508.01390}, 
}

@misc{han2026socialphysicsageartificial,
      title={Social physics in the age of artificial intelligence}, 
      author={The Anh Han and Joel Z. Leibo and Tom Lenaerts and Iyad Rahwan and Fernando Santos and Matjaž Perc and Valerio Capraro},
      year={2026},
      eprint={2603.16900},
      archivePrefix={arXiv},
      primaryClass={physics.soc-ph},
      url={https://arxiv.org/abs/2603.16900}, 
}

@article{sakai2026effects,
  title   = {Effects of personality steering on cooperative behavior in large language model agents},
  author  = {Sakai, M. and Yokoyama, M. and Tateishi, W. and Ichinose, G},
  journal = {Scientific Reports},
  year    = {2026},
  doi     = {10.1038/s41598-026-56163-8},
  publisher = {Nature Publishing Group}
}

@inproceedings{ren2024emergence,
author = {Ren, Siyue and Cui, Zhiyao and Song, Ruiqi and Wang, Zhen and Hu, Shuyue},
title = {Emergence of social norms in generative agent societies: principles and architecture},
year = {2024},
isbn = {978-1-956792-04-1},
url = {https://doi.org/10.24963/ijcai.2024/874},
doi = {10.24963/ijcai.2024/874},
booktitle = {Proceedings of the Thirty-Third International Joint Conference on Artificial Intelligence},
articleno = {874},
numpages = {9},
location = {Jeju, Korea},
series = {IJCAI '24}
}

@article{Argyle_Busby_Fulda_Gubler_Rytting_Wingate_2023, title={Out of One, Many: Using Language Models to Simulate Human Samples}, volume={31}, DOI={10.1017/pan.2023.2}, number={3}, journal={Political Analysis}, author={Argyle, Lisa P. and Busby, Ethan C. and Fulda, Nancy and Gubler, Joshua R. and Rytting, Christopher and Wingate, David}, year={2023}, pages={337–351}}

@inproceedings{jiang-etal-2024-personallm,
    title = "{P}ersona{LLM}: Investigating the Ability of Large Language Models to Express Personality Traits",
    author = "Jiang, Hang  and
      Zhang, Xiajie  and
      Cao, Xubo  and
      Breazeal, Cynthia  and
      Roy, Deb  and
      Kabbara, Jad",
    editor = "Duh, Kevin  and
      Gomez, Helena  and
      Bethard, Steven",
    booktitle = "Findings of the Association for Computational Linguistics: NAACL 2024",
    month = jun,
    year = "2024",
    address = "Mexico City, Mexico",
    publisher = "Association for Computational Linguistics",
    url = "https://aclanthology.org/2024.findings-naacl.229/",
    doi = "10.18653/v1/2024.findings-naacl.229",
    pages = "3605--3627"
}

@inproceedings{jiang2023evaluating,
author = {Jiang, Guangyuan and Xu, Manjie and Zhu, Song-Chun and Han, Wenjuan and Zhang, Chi and Zhu, Yixin},
title = {Evaluating and inducing personality in pre-trained language models},
year = {2023},
publisher = {Curran Associates Inc.},
address = {Red Hook, NY, USA},
booktitle = {Proceedings of the 37th International Conference on Neural Information Processing Systems},
articleno = {466},
numpages = {22},
location = {New Orleans, LA, USA},
series = {NIPS '23}
}

@article{Holt_Sullivan_2023, title={Permutation tests for experimental data}, volume={26}, DOI={10.1007/s10683-023-09799-6}, number={4}, journal={Experimental Economics}, author={Holt, Charles A. and Sullivan, Sean P.}, year={2023}, pages={775–812}}

@inproceedings{pan-etal-2023-risk,
    title = "On the Risk of Misinformation Pollution with Large Language Models",
    author = "Pan, Yikang  and
      Pan, Liangming  and
      Chen, Wenhu  and
      Nakov, Preslav  and
      Kan, Min-Yen  and
      Wang, William",
    editor = "Bouamor, Houda  and
      Pino, Juan  and
      Bali, Kalika",
    booktitle = "Findings of the Association for Computational Linguistics: EMNLP 2023",
    month = dec,
    year = "2023",
    address = "Singapore",
    publisher = "Association for Computational Linguistics",
    url = "https://aclanthology.org/2023.findings-emnlp.97/",
    doi = "10.18653/v1/2023.findings-emnlp.97",
    pages = "1389--1403"
}

@inproceedings{chen2024can,
  title     = {Can {LLM}-Generated Misinformation Be Detected?},
  author    = {Chen, Canyu and Shu, Kai},
  booktitle = {Proceedings of the Twelfth International Conference on Learning Representations (ICLR)},
  year      = {2024}
}

@article{gao2025takecaution,
author = {Yuan Gao  and Dokyun Lee  and Gordon Burtch  and Sina Fazelpour },
title = {Take caution in using LLMs as human surrogates},
journal = {Proceedings of the National Academy of Sciences},
volume = {122},
number = {24},
pages = {e2501660122},
year = {2025},
doi = {10.1073/pnas.2501660122},
URL = {https://www.pnas.org/doi/abs/10.1073/pnas.2501660122},
eprint = {https://www.pnas.org/doi/pdf/10.1073/pnas.2501660122}}

@inproceedings{rieger2026ChatGPT,
author = {Rieger, Tobias and Roesler, Eileen},
title = {ChatGPT Is a People Pleaser, Especially When Taking Human vs. AI Advice},
year = {2026},
isbn = {9798400722813},
publisher = {Association for Computing Machinery},
address = {New York, NY, USA},
url = {https://doi.org/10.1145/3772363.3798923},
doi = {10.1145/3772363.3798923},
booktitle = {Proceedings of the Extended Abstracts of the 2026 CHI Conference on Human Factors in Computing Systems},
articleno = {161},
numpages = {6},
location = {
},
series = {CHI EA '26}
}

@misc{qwen3.5,
    title  = {{Qwen3.5}: Towards Native Multimodal Agents},
    author = {{Qwen Team}},
    month  = {February},
    year   = {2026},
    url    = {https://qwen.ai/blog?id=qwen3.5}
}

@misc{gemmateam2026gemma4,
      title={Gemma 4 Technical Report}, 
      author={Gemma Team},
      year={2026},
      eprint={2607.02770},
      archivePrefix={arXiv},
      primaryClass={cs.CL},
      url={https://arxiv.org/abs/2607.02770}, 
}
}

\clearpage
\newpage

\noindent
\textbf{Data, Materials, and Software Availability.} The data we collected from human participants on Prolific, our LLM simulation setup, LLM data and the analysis code are available on Github:
\begin{quote}
    https://github.com/oneSebastian/BeliefUpdateSimulation.
\end{quote}
The dataset is also released independently at Huggingface:
\begin{quote}
    https://huggingface.co/datasets/SebastianPohl/BeliefUpdateSimulation.
\end{quote}

\noindent
\textbf{Acknowledgments.} Christian Hilbe acknowledges generous support by the European Research Council grant 850529: E-Direct. We would further like to thank Theodoros Saroglou, Sebastian Dennerlein, Philipp Wintersberger, Bernd Resch and Daniel Klotz for helpful discussions of the problems addressed in this paper.\\

\noindent
\textbf{Author contributions.}
SP, HM, PM, AG, FL jointly conceived the research question and experiments. SP: manuscript writing and additional experiments, HM: mixed effects model analysis and manuscript writing, PM: statistical analysis, AG: LLM data collection, FL: human data collection. YH and CH jointly supervised the work and advised on experiment design, data analysis and manuscript writing. \\

\noindent
\textbf{Competing interests}
The authors declare no competing interests.

\clearpage
\newpage

\end{document}